\documentclass[10pt,twocolumn,letterpaper]{article}

\usepackage{iccv}              
\usepackage{algorithm}
\usepackage[noend]{algpseudocode}
\usepackage{makecell} 
\usepackage{multirow}
\newcommand*{\affaddr}[1]{#1} 
\newcommand*{\affmark}[1][*]
{\textsuperscript{#1}}

\usepackage{pifont}
\usepackage[pagebackref,breaklinks,colorlinks,allcolors=iccvblue]{hyperref}

\newcommand*{\belowrulesepcolor}[1]{%
  \noalign{%
    \kern-\belowrulesep 
    \begingroup 
      \color{#1}%
      \hrule height\belowrulesep 
    \endgroup 
  }%
} 
\newcommand*{\aboverulesepcolor}[1]{%
  \noalign{%
    \begingroup 
      \color{#1}%
      \hrule height\aboverulesep 
    \endgroup 
    \kern-\aboverulesep 
  }%
}

\definecolor{myblue}{RGB}{230,230,255}

\definecolor{iccvblue}{rgb}{0.21,0.49,0.74}
\definecolor{fullgreen}{rgb}{0.502, 0.788, 0.643}
\definecolor{fullred}{rgb}{0.800, 0.447, 0.541}
\definecolor{fullwhite}{rgb}{225, 225, 225}
\newcommand{\cmark}{\textcolor{fullgreen}{\textbf{\checkmark}}}
\newcommand{\xmark}{\textcolor{fullred}{\textbf{\ding{55}}}}

\usepackage{color}      
\usepackage{colortbl}   

\def\paperID{4702} 
\def\confName{ICCV}
\def\confYear{2025}

\title{ForCenNet: Foreground-Centric Network for Document Image Rectification}

\author{
Peng Cai\affmark[1], 
Qiang Li\affmark[1], 
Kaicheng Yang\affmark[2], 
Dong Guo\affmark[1], 
Jia Li\affmark[1], 
Nan Zhou\affmark[1] \\
Xiang An\affmark[2],  
Ninghua Yang\affmark[2],  
Jiankang Deng\affmark[3]\thanks{Corresponding Author}\\
\affaddr{\affmark[1]Qihoo Technology} \qquad
\affaddr{\affmark[2]DeepGlint} \qquad
\affaddr{\affmark[3]Imperial College London}\\
{\tt\small \{caipeng1,liqiang-s\}@360.cn}
}

\begin{document}
\maketitle

\begin{abstract}
Document image rectification aims to eliminate geometric deformation in photographed documents to facilitate text recognition. However, existing methods often neglect the significance of foreground elements, which provide essential geometric references and layout information for document image correction. In this paper, we introduce \textbf{For}eground-\textbf{Cen}tric \textbf{Net}work~(\textbf{ForCenNet}) to eliminate geometric distortions in document images. Specifically, we initially propose a foreground-centric label generation method, which extracts detailed foreground elements from an undistorted image. Then we introduce a foreground-centric mask mechanism to enhance the distinction between readable and background regions. Furthermore, we design a curvature consistency loss to leverage the detailed foreground labels to help the model understand the distorted geometric distribution. Extensive experiments demonstrate that ForCenNet achieves new state-of-the-art on four real-world benchmarks, such as DocUNet, DIR300, WarpDoc, and DocReal. Quantitative analysis shows that the proposed method effectively undistorts layout elements, such as text lines and table borders. The resources for further comparison are provided at~\url{https://github.com/caipeng328/ForCenNet}.
\end{abstract}
    
\section{Introduction}

With the widespread adoption of mobile devices and advancements in camera technology, document digitization has become essential. Smartphones and portable cameras, offering greater flexibility than traditional scanners, often capture images that suffer from perspective distortion and geometric deformation due to variable shooting angles and document conditions. These distortions negatively impact visual quality and hinder downstream tasks such as OCR and document structure analysis. Therefore, accurate rectification of documents captured by camera sensors remains a significant challenge in the field of document analysis and recognition.

With advances in deep learning, data-driven approaches~\cite{das2019dewarpnet, feng2021doctr, verhoeven2023uvdoc} have been developed for geometric correction of distorted document images. Unlike traditional computer vision tasks such as image classification~\cite{huang2017densely} and object detection~\cite{he2017mask}, acquiring fine-grained annotations for document dewarping poses significant challenges. The limited availability of training data restricts the models' generalization capabilities. Certain methods~\cite{ma2022learning,xue2022fourier,liu2023rethinking} introduce random perturbations to distorted images and assess the consistency between the corrected images and synthetic deformation patterns. While these weakly-supervised methods achieve satisfactory results in terms of image similarity when compared to contemporary studies, they often fail to preserve document readability. This observation highlights the need to reassess the data scale and motivates the exploration of more diverse synthetic deformation fields to better approximate the true distribution.

\begin{figure}[t]
\centering
\includegraphics[width=\linewidth]{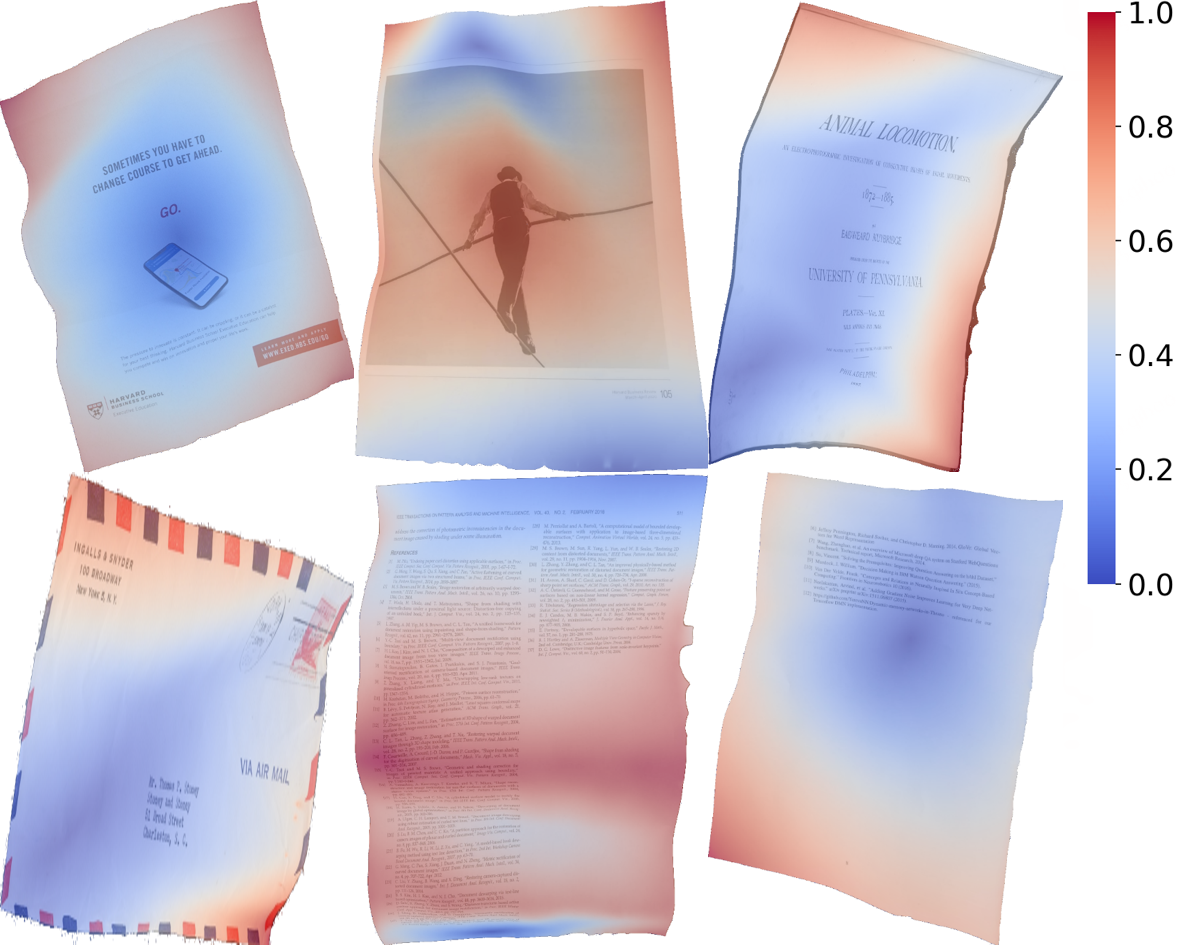}
\caption{\textbf{Visualization of Deformation Degree.} We compute the deformation displacement for each pixel and visualize it using a heatmap. Blue represents minor deformation, while red indicates significant displacement.}
\label{fig:image1}
\vspace{-5mm}
\end{figure}

The accuracy of information extraction and content readability are fundamental objectives in document dewarping. As shown in Fig.~\ref{fig:image1}, readable regions in the document, such as text and table lines, occupy only a small fraction of the image pixels, while major distortions are primarily concentrated in the background. Current methods~\cite{zhang2024docres,verhoeven2023uvdoc,feng2021doctr,zhang2022marior,bandyopadhyay2021gated} uniformly predict warping flows across the entire image, leading to a misalignment between the task’s primary objectives and the model’s optimization goals. This misalignment complicates the optimization process. Consequently, the precise definition of foreground regions in document dewarping has emerged as a critical research focus. Recent approaches, such as DocGeoNet~\cite{feng2022geometric} and FTDR~\cite{li2023foreground}, emphasize text lines as foreground elements to guide model predictions. DocGeoNet~\cite{feng2022geometric} incorporates mask annotations for the Doc3D~\cite{das2019dewarpnet} dataset, while FTDR~\cite{li2023foreground} utilizes a frozen detection model to extract coarse text line information from distorted images. However, these methods exhibit two notable limitations: (1) conventional detection models~\cite{ronneberger2015u,zhong2019publaynet} struggle to accurately identify distorted text lines and linear structures in geometrically deformed document images, and (2) readable information in documents encompasses not only text lines but also additional foreground elements, such as table lines and graphical content.

To address the aforementioned limitations, this paper introduces the \textbf{For}eground-\textbf{Cen}tric \textbf{Net}work~(\textbf{ForCenNet}) for eliminating geometric distortions in document images. Specifically, we propose a foreground-centric label generation method to extract detailed foreground elements, including text, straight lines, and drawings, from undistorted images. These foreground elements are represented using both masks and discrete points. The undistorted image, along with its corresponding mask and point annotations, undergoes a forward mapping process, while the backward mapping serves as the prediction target for the model. Given that foreground regions encapsulate the most significant distortion magnitudes in dewarped images, we introduce a foreground-centric mask mechanism to enhance the differentiation between readable and background regions. Additionally, a curvature consistency loss is designed to exploit detailed foreground labels, allowing the model to better capture the geometric structure of distortions. Extensive experiments demonstrate that ForCenNet achieves state-of-the-art performance on four real-world benchmarks: DocUNet, DIR300, WarpDoc, and DocReal. The main contributions of this paper are summarized as follows:
\begin{itemize}[noitemsep,topsep=0pt]
\item We propose a foreground-centric label generation method to address data scarcity in document rectification. This module generates and utilizes intrinsic foreground information, enabling efficient training with only distortion-free reference images.
\item We introduce a foreground-centric mask mechanism to improve the distinction between readable content and background regions, effectively capturing primary distortion patterns in dewarped images.
\item We design a curvature consistency loss to leverage detailed foreground labels, enhancing the model's ability to capture the geometric distribution of distortions.
\item We conduct extensive experiments and demonstrate that the proposed ForCenNet establishes a new state-of-the-art across four real-world benchmarks.
\end{itemize}
\section{Related Work}
\subsection{Deformation Field Supervised for Rectification}
The advancement of deep learning techniques~\cite{Vaswani_Shazeer_Parmar_Uszkoreit_Jones_Gomez_Kaiser_Polosukhin_2017, wang2021pyramid,gu2025breaking,yang2025clip,an2024multi} has established deep networks~\cite{Teed_Deng_2021} as an effective alternative to traditional methods to predict pixel-level deformation fields. Compared to conventional approaches~\cite{Brown_Seales_2002, Huaigu_Cao_XiaoqingDing_ChangsongLiu_2004}, deep learning-based document image dewarping methods~\cite{ma2018docunet, Li_Zhang_Liao_Sander_2019, feng2021doctr, zhang2022marior, verhoeven2023uvdoc, hertlein2023template,feng2021docscanner} achieve superior feature representation. DocUNet~\cite{ma2018docunet} utilizes a stacked U-Net~\cite{ronneberger2015u} to estimate pixel-wise displacement fields for document distortion correction. PWUNet~\cite{das2021end} introduces an end-to-end trainable piecewise dewarping framework that integrates local deformation prediction with global structural constraints. DocTr~\cite{feng2021doctr} employs transformer-based architectures~\cite{Vaswani_Shazeer_Parmar_Uszkoreit_Jones_Gomez_Kaiser_Polosukhin_2017} to enhance representation learning and correction accuracy. Marior~\cite{zhang2022marior} addresses large-margin document distortions by incorporating edge-removal modules and content-aware loss functions. CGU-Net~\cite{verhoeven2023uvdoc} predicts 3D and 2D distorted grids using a fully convolutional network, effectively modeling the transformation from 3D to 2D. To alleviate data scarcity in document dewarping, recent methods~\cite{ma2022learning, xue2022fourier, liu2023rethinking} adopt weak supervision strategies. PaperEdge~\cite{ma2022learning} fine-tunes the contour network ENet using the weakly supervised mask-labeled dataset DIW~\cite{ma2022learning}, enabling self-supervised learning through its texture prediction network. FDRNet~\cite{xue2022fourier} leverages WarpDoc~\cite{xue2022fourier} to generate dual distortion modes and predicts control points with shared parameter weights for mutual correction.  DRNet~\cite{liu2023rethinking} introduces a novel supervision approach using undistorted images as direct supervisory signals. In contrast to these methods, our approach facilitates rapid model iteration using only undistorted images, thereby supporting the development of an efficient and cost-effective training framework.

\subsection{Foreground Constraints Based Rectification}
Several studies~\cite{ulges2005document, lavialle2001active, xie2021document, feng2022geometric, jiang2022revisiting, li2023foreground, li2023layout, meng2011metric, wu2002document} focus on utilizing inherent foreground constraints in document images to optimize geometric rectification. Early works~\cite{ulges2005document, meng2011metric, wu2002document} employ polynomial approximation to model curved text lines and correct document images captured by single cameras, relying on a priori layout information. Similarly, \cite{lavialle2001active} utilizes cubic B-splines~\cite{prautzsch2002bezier} and an active contour network, to improve the accuracy of straightening curved text lines. Recent methods, such as DDCP~\cite{xie2021document}, predict a fixed set of foreground control points and compute backward mapping based on their relationship to reference points. DocGeoNet~\cite{feng2022geometric} enhances correction performance by integrating segmentation loss to guide a CNN-based text line extractor. RDGR~\cite{jiang2022revisiting} identifies text lines and boundaries, followed by grid regularization for backward mapping, ensuring structural preservation during dewarping. FTDR~\cite{li2023foreground} employs global-local fusion and cross-attention mechanisms to emphasize foreground and text line regions, improving readability and correction quality. LA-DocFlatten~\cite{li2023layout} introduces a transformer-based segmentation module to capture foreground layout, combined with regression and merging modules for UV mapping prediction. In this work, we explicitly define foreground elements, including text, lines, and vector graphics, and design curvature consistency loss and mask-guided mechanisms to enhance the model’s geometric understanding of foreground elements.

\begin{figure*}[ht!]
\centering
\begin{subfigure}{0.28\linewidth}
\includegraphics[width=1.0\linewidth]{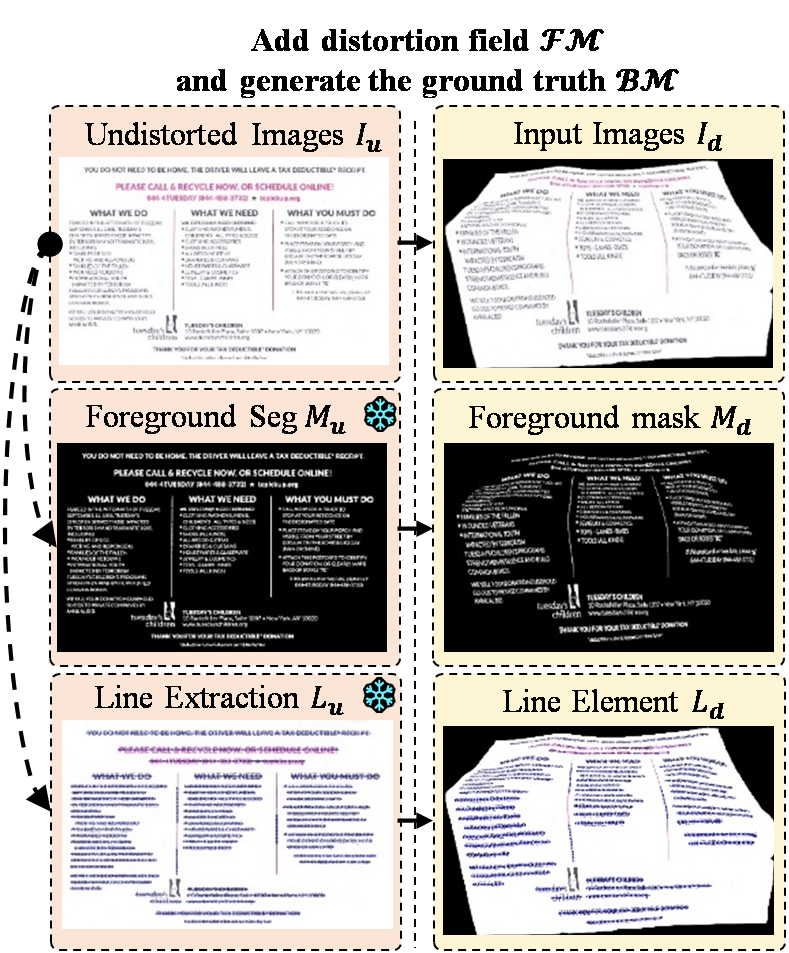}
\subcaption[]{Label Generation}
\label{fig:image3_1}
\end{subfigure}
\hspace{2mm}
\begin{subfigure}{0.4\linewidth}
\includegraphics[width=1.0\linewidth]{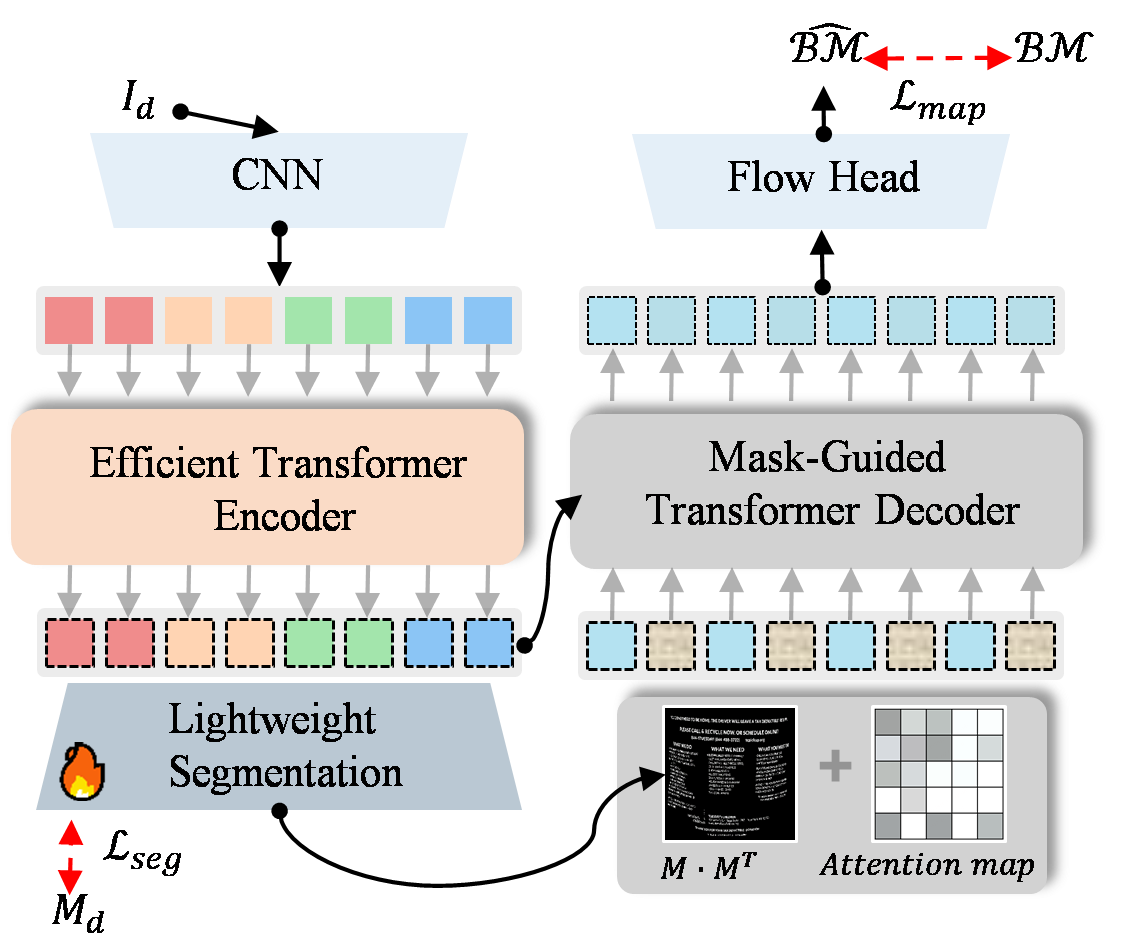}
\subcaption[]{The Architecture of ForCenNet}
\label{fig:image3_2}
\end{subfigure}
\hspace{2mm}
\begin{subfigure}{0.23\linewidth}
\includegraphics[width=1.0\linewidth]{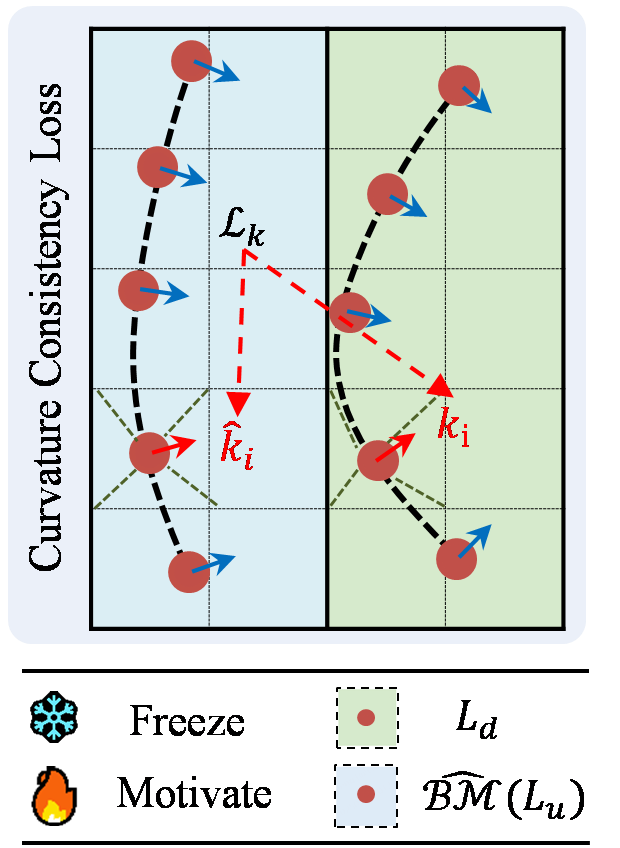}
\subcaption[]{Curvature Consistency Loss}
\label{fig:image3_3}
\end{subfigure}

\vspace{-2mm}
\caption{\textbf{The overview architecture of the proposed ForCenNet.} $M$ is the predicted foreground mask. $k$ is the curvature value calculated from line elements, with $\hat{k_i}$ as the predicted value and $k_i$ as the ground truth. $\mathcal{\hat{BM}}$ is the predicted backward mapping field.}
\label{fig:causal}
\vspace{-3mm}
\end{figure*}

\section{Methods}
\indent In this section, we propose ForCenNet, a novel framework designed for efficient document rectification using only undistorted images. As illustrated in Fig.~\ref{fig:causal}, foreground elements are first extracted from undistorted images (Section~\ref{3.1:labelgeneration}), then enhance the model's focus on the foreground through a mask-guided module (Section~\ref{3.2:NetworkArchitecture}), and finally design a curvature consistency loss (Sec.~\ref{3.3OptimizationObjectives})  to improve the model's geometric perception of small elements, such as table lines. 

\subsection{Foreground-Centric Label Generation}
\label{3.1:labelgeneration}

We introduce a foreground-centric label generation method to alleviate data scarcity in document correction tasks. To address inaccuracies caused by image deformation and varying lighting conditions, our approach extracts precise foreground elements from undistorted images and applies a randomly generated distortion field to generate accurate training samples. As shown in Fig.~\ref{fig:image3_1}, the proposed method operates in three stages: foreground-background segmentation, line element extraction, and distortion field generation.

\noindent{\bf Character-Level Foreground-Background Segmentation.} 
To obtain character-level foreground-background segmentation, we fine-tune Hi-SAM~\cite{ye2025hi} for segmenting readable regions in document images. As shown in Fig.~\ref{fig:image3_1}, given an undistorted input image $I_u$, it generates the corresponding foreground mask $M_u$. The fine-tuned Hi-SAM unifies the segmentation of text regions, line elements, and graphics, which will serve as subsequent supervision and guidance signals.

\begin{algorithm}[t!]
\caption{Line Segment Filtering}
\textbf{Require:} Undistorted image $I$, slope threshold $\epsilon_{s}$, intercept \par 
\hspace{1em} threshold $\delta$, slope ranges $(0, \alpha)$ for horizontal lines and\par 
\hspace{1em} $(\beta, \infty)$ for vertical lines.
\begin{algorithmic}[1]
    \State $edges \gets \text{Canny}(I)$
    \State $line\_segments \gets \text{LSD}(edges)$
    \For{line \textbf{in} line\_segments}
        \State $(x_0, y_0), (x_1, y_1) \gets line$
        \State $slope \gets \left| \frac{y_1 - y_0}{x_1 - x_0} \right|$
        \If{$slope < \alpha$ \textbf{or} $slope > \beta$}
            \State \textbf{keep}(line)
        \EndIf
    \EndFor
    \For{line1, line2 \textbf{in} line\_segments}
        \State $(m_1, c_1) \gets \text{calculate\_slope\_and\_intercept}(line1)$
        \State $(m_2, c_2) \gets \text{calculate\_slope\_and\_intercept}(line2)$
        \If{$|m_1 - m_2| < \epsilon_{s}$ \textbf{and} $|c_1 - c_2| < \delta$}
            \State \textbf{remove}(line2)
        \EndIf
    \EndFor 
\State \textbf{Return} line\_segments
\end{algorithmic}
\label{algorithm}
\end{algorithm}

\noindent{\bf Extraction of Line Elements.} 
We employ OCR engines~\cite{PaddleOCR} to extract text lines, utilizing the midlines of text bounding boxes as the text line representations. For document line elements such as table lines, we introduce a document-specific detection method based on the Line Segment Detector (LSD)~\cite{von2008lsd}, which eliminates non-horizontal and non-vertical lines and suppresses duplicate detections. The implementation details are provided in Algorithm~\ref{algorithm}.

\noindent{\bf Distortion Field Generation.} 
We obtain the native backward mapping $\mathcal{BM}$ from DOC3D~\cite{das2019dewarpnet} to compute the corresponding forward mapping $\mathcal{FM}$. To enrich the deformation field, we slightly crop and apply random pairwise overlapping to the $\mathcal{BM}$. Subsequently, the $\mathcal{FM}$ is superimposed onto the undistorted image, foreground mask, and line elements to generate the distorted image $I_d$, foreground mask $M_{d}$, and line elements $L_{d}$, with $\mathcal{BM}$ serving as the training target.

\subsection{Foreground-Centric Network Architecture}
\label{3.2:NetworkArchitecture}

Existing methods~\cite{verhoeven2023uvdoc,zhang2024docres} often treat text, line elements, and background uniformly, overlooking the significance of foreground regions. To address this, we employ distortion masks to guide the model in focusing on foreground information, prioritizing text readability and structural alignment. As shown in Fig.~\ref{fig:image3_2}, our proposed ForCenNet comprises four key components: Feature Extraction Module, Efficient Transformer Encoder, Foreground Segmentation Module, and Mask-Guided Transformer Decoder.

\noindent{\bf Feature Extraction Module.} 
Given a distorted document image $I_d \in \mathbb{R}^{H \times W \times C}$, we resize it to $H=W=288$ and $C=3$. The resized image is passed through convolutional layers with large kernels and multiple residual layers, yielding the features $F_u \in \mathbb{R}^{\frac{H}{8} \times \frac{W}{8} \times 256}$.

\noindent{\bf Efficient Transformer Encoder.} To capture global dependencies, we utilize a vanilla Transformer~\cite{Vaswani_Shazeer_Parmar_Uszkoreit_Jones_Gomez_Kaiser_Polosukhin_2017} with three layers. We adopt overlapping patch embeddings~\cite{touvron2021training} 
and employ a kernel size of 3 and a stride of 2 to preserve feature information at text boundaries while reducing computational overhead. The Transformer encoder produces three feature maps: $\{E_1, E_2, E_3\}$. To reduce the complexity of the attention mechanism, we implement a \textbf{S}patial \textbf{P}ooling \textbf{W}indow (SPW) strategy~\cite{jia2024transformer} on the key and value features.

\noindent{\bf Foreground Segmentation Model.} 
Given the feature sequence $E$, the foreground segmentation model predicts a binary mask using a lightweight network. First, the feature channel dimensions are unified via a 1×1 convolution, followed by upsampling to a spatial resolution of $H \times W$. The merged features are then processed through multiple 1×1 convolutions to produce the foreground segmentation result $M \in \mathbb{R}^{H \times W \times C_{seg}}$, where $C_{seg}=2$ represents the foreground and background classes. The training process is supervised using an L1 loss with the ground truth provided by $M_{d}$:
{\small
\begin{equation}
\label{L_seg}
\mathcal{L}_{seg} = ||M-M_{d}||_1.
\end{equation}
}
To incorporate the foreground segmentation result $M$ into the decoder, we apply a softmax operation with a smoothing coefficient to $M$, yielding pixel-wise class probabilities. The expected value of these probabilities is then computed to generate the predicted foreground mask $\tilde{M}$:
{\small
\begin{equation}
\tilde{M} = \sum_{i \in \{0,1\}} i \cdot softmax( \gamma \cdot M)_i,
\end{equation}
}
where $\gamma$ denotes the smoothing coefficient. The softmax operation normalizes $M$, ensuring that its values sum to 1, which allows the expected value to be interpreted as a probability density.

\noindent{\bf Mask-Guided Transformer Decoder.} We use the foreground mask $\tilde{M}$ to guide feature extraction in the Transformer decoder. The decoder takes the feature sequence $\{E_1, E_2, E_3\}$ and the foreground mask $\tilde{M}$ as inputs, utilizing learnable embeddings $Q_{learn}$ to capture distortion information, and outputs the distorted deformation field. The encoder consists of three Transformer layers, with upsampling to connect them. Each transformer layer includes mask-guided self-attention and encoder-decoder cross-attention mechanisms. For mask-guided self-attention, we use decoder embeddings $D_I$ as input, incorporating the foreground mask $\tilde{M}$ to focus attention on foreground regions. In encoder-decoder cross-attention, the decoder embeddings serve as queries, and the encoder embeddings from the same layer act as keys and values.
{\small
\begin{equation}
\begin{aligned}
\mathrm{MSA}(Q, K, V) &= \mathrm{Softmax}(\frac{QK^T + \sigma Seq(\tilde{M})Seq(\tilde{M})^T}{\sqrt{d_{\mathrm{head}}}})V, \\
\mathrm{CA}(Q, K, V) &= \mathrm{CrossAttention}(D_{i-1}, E_i, E_i),
\end{aligned}
\end{equation}
}
where MSA denotes mask-guided self-attention, with $\sigma$ as the scaling factor. ``Seq.'' refers to sequential expansion along the feature dimension. ``CA'' represents encoder-decoder cross-attention, where $D_{i-1}$ and $E_i$ are the outputs of the $(i-1)$-th decoder layer and $i$-th encoder layer, respectively. To reduce attention complexity, spatial reduction~\cite{jia2024transformer} is applied to the key and value in the decoder. Finally, we use the upsampling method from DocTr~\cite{feng2021doctr} and DocGeoNet~\cite{feng2022geometric} to obtain a high-resolution backward deformation field $\mathcal{BM}$, which initializes grid coordinates and refines them through weighted optimization with learnable parameters.

\begin{figure*}[t!]
\centering\includegraphics[width=0.88\linewidth]{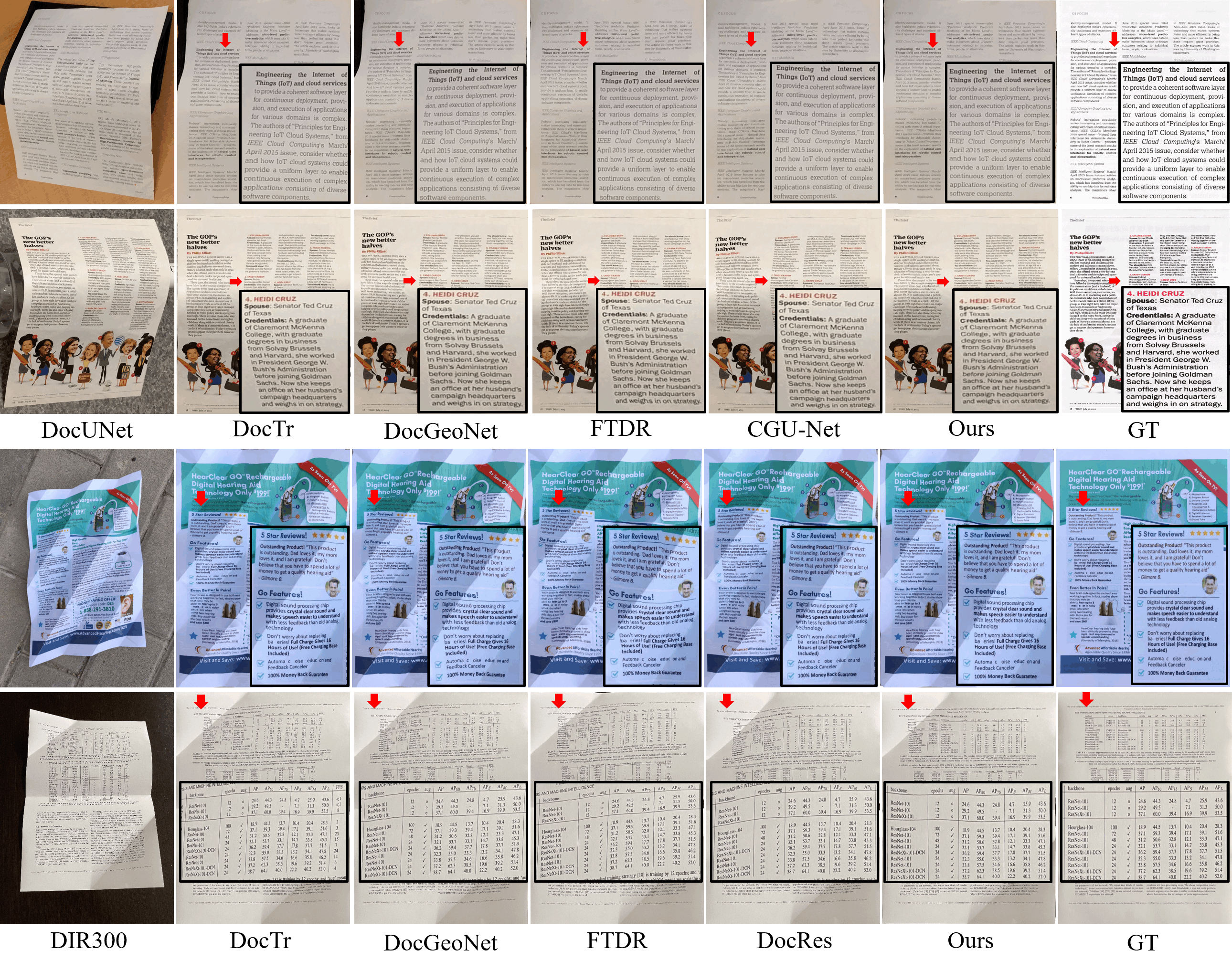}
\vspace{-2mm}
\caption{\textbf{Qualitative Comparison with Prior Methods on DocUNet and DIR300 Benchmarks.} Red arrows highlight the differences. Additional visualizations are available in the appendix.}
\vspace{-3mm}
\label{fig:image7}
\end{figure*}

\subsection{Foreground-Centric Optimization Objectives}
\label{3.3OptimizationObjectives}

The ForCenNet has three objectives: foreground mask loss $\mathcal{L}_{seg}$, backward map regression $\mathcal{L}_{map}$, and curvature consistency loss $\mathcal{L}_{k}$. $\mathcal{L}_{seg}$ supervises the generation of the foreground mask, which is defined as Eq.~\ref{L_seg}. Backward map regression loss is calculated by using the L1 distance between the predicted $\hat{\mathcal{BM}}$ and the ground truth $\mathcal{BM}$, as defined by the formula: $\mathcal{L}_{map} = ||\hat{\mathcal{BM}} - \mathcal{BM}||_1$.

\noindent{\bf Curvature Consistency Loss.} Compared to foreground elements like text and images, table lines occupy fewer pixels, making it challenging for the network to capture line distortion. This imbalance weakens the supervisory effect of the L1 loss. Furthermore, although L1 loss governs pixel-level distortion offsets, it does not adequately capture the geometric structure of linear elements. To resolve these issues, we introduce a curvature consistency loss based on control lines, utilizing line elements $L_{u}$ from undistorted images.

As illustrated Fig.~\ref{fig:image3_1}, 
every 4 pixels along each extracted line element $L_{u}$ are sampled to generate a point set $P = \{ p_i \mid p_i \in L_{u}, i=1, 2, \ldots, N \}$. This set is projected onto the predicted deformation field $\hat{\mathcal{BM}}$ and the ground truth deformation field $\mathcal{BM}$ to obtain the predicted control points $Cp$ and the ground truth control points $Cp_{gt}$. To minimize errors and ensure differentiability, bilinear interpolation~\cite{kirkland2010bilinear} is applied for point mapping and projection, formulated as:
{\small
\begin{equation}
\begin{aligned}
Cp &= \{\sum_{p \in N(p_i)} w_p \cdot \hat{BM}(p) \mid p_i \in P \}, \\ 
Cp_{gt} &= \{\sum_{p \in N(p_i)} w_p \cdot BM(p) \mid p_i \in P \},
\end{aligned}
\end{equation}
}
where $w_{(p)}$ is the bilinear interpolation weight determined by the relative position of $p_i$ on the grid. $N(p_i)$ denotes the set of four neighboring points surrounding $p_i$. For the mapped point set $Cp = \{(x_i, y_i) \mid i = 1, 2, \ldots, N \}$, the derivatives are calculated using the central difference method~\cite{chapra2011numerical}, while the forward and backward differences are applied at the boundary points. The curvature of the control line is given by:
{\small
\begin{equation}
\kappa_i = \frac{|x_i' \times y_i'' - y_i'  \times x_i''|}{(x_i'^2 + y_i'^2)^{3/2} + \varepsilon},
\end{equation}
}
where $x_i'$, $x_i''$, $y_i'$, and $y_i''$ represent the first and second discrete derivatives of the coordinates. To prevent gradient explosion or overflow, a small positive value $\varepsilon=0.0001$ is added. Finally, a constraint is imposed to ensure that the curvature variation at local points aligns with the true curvature trend. The loss function $\mathcal{L}_{k}$ is defined as:
{\small
\begin{equation}
\mathcal{L}_{k} = \frac{1}{N-1} \sum_{i}^{N-1}(\hat{k_i} - k_i),
\end{equation}
}
where $\hat{k_i}$ and $k_i$ represent the curvature values at discrete points $P$ in the predicted and ground truth deformation fields.

\begin{table}[t]
\centering

\resizebox{\linewidth}{!}{
\begin{tabular}{llccccc}
\toprule
\textbf{Type} & \textbf{Model} & \textbf{MS-SSIM$\uparrow$} & \textbf{LD$\downarrow$} & \textbf{AD$\downarrow$} & \textbf{ED$\downarrow$} & \textbf{CER$\downarrow$} \\ 
\midrule
\multirow{3}{*}{WS. }  
    & PaperEdge~\cite{ma2022learning}       & 0.470  & 8.49  & 0.39 & 825.48 & 0.211  \\ 
    & FDRNet~\cite{xue2022fourier}  & 0.542  & 8.21  & --         & 829.78 & 0.206  \\ 
    & DRNet~\cite{liu2023rethinking}    & 0.510  & 7.42  & --         & 644.48 & 0.164   \\ 
\midrule
\multirow{5}{*}{Other.} 
    & DewarpNet~\cite{das2019dewarpnet}       & 0.473  & 8.39  & 0.42  & 885.90 & 0.237 \\
    & PWUNet~\cite{das2021end}      & 0.491
    & 8.64  & --  & 1069.28 & 0.267 \\
    & DocTr~\cite{feng2021doctr}      & 0.510
    & 7.75 & 0.39 & 724.84 & 0.183 \\
    & Marior~\cite{zhang2022marior}      & 0.478  & 7.43  & 0.40  & 823.80 & 0.205 \\
    & CGU-Net~\cite{verhoeven2023uvdoc} & 0.557  & 6.83  & 0.31 & \textbf{513.76} & 0.178 \\
\midrule
\multirow{5}{*}{FU.} 
    & DDCP~\cite{xie2021document}    & 0.472  & 8.97          & 0.45 & 1411.38  & 0.357  \\ 
    & DocGeoNet~\cite{feng2022geometric}          & 0.504 & 7.71  & 0.38          & 713.94 & 0.182 \\ 
    & RDGR~\cite{jiang2022revisiting}       & 0.495  & 8.51  & 0.46  & 729.52 & 0.171 \\
    & FTDR~\cite{li2023foreground}          & 0.497  & 8.43  & 0.37            & 697.52   & 0.170 \\ 
    & LA-DocFlatten~\cite{li2023layout}      & 0.526  & 6.72  & 0.30 & -- & -- \\
    \rowcolor{myblue} 
    {\cellcolor{fullwhite}}
    &ForCenNet-DOC3D  & \underline{ 0.579} & \underline{4.91} &\underline{0.19}  & 592.37
    &\underline{ 0.158}\\ 
    \rowcolor{myblue} 
    {\cellcolor{fullwhite}}
    &ForCenNet  & \textbf{0.582} & \textbf{4.82} &\textbf{0.19}  & \underline{571.40}
    &\textbf{0.136}\\ 

\bottomrule
\end{tabular}
}
\caption{\textbf{Result comparisons} between  our proposed with existing methods on the DocUNet Benchmark~\cite{ma2018docunet}. \textbf{WS.} refers to weakly supervised methods. \textbf{FU.} refers to methods that leverage foreground elements. Bolded values indicate the best, and underlined values indicate the second best.}
\label{tbl:docunet}
\vspace{-3mm}
\end{table}

\section{Experiments}

\subsection{Implementation Details}
We implement ForCenNet using the PyTorch framework~\cite{paszke2017automatic}. We train our model on two distinct undistorted datasets. The first, referred to as ForCenNet, comprises 365 images from DocUNet~\cite{ma2018docunet} and DIR300~\cite{feng2022geometric}. The second variant, ForCenNet-DOC3D, is trained on undistorted images from the DOC3D dataset~\cite{das2019dewarpnet}. The initial backward mapping field is generated from 100,000 samples of the Doc3D dataset~\cite{das2019dewarpnet}. To simulate real-world conditions, we overlay distorted document images onto random COCO backgrounds~\cite{lin2014microsoft}. The training images are resized to $288\times288$ pixels. Optimization is performed using the AdamW optimizer~\cite{Loshchilov_Hutter_2017} with a batch size of 32, and the OneCycle learning rate scheduler~\cite{Smith_Topin_2019} is employed, setting the maximum learning rate at $10^{-4}$. The warm-up phase~\cite{goyal2017accurate} constitutes 10\% of the total training cycles. We train for 30 epochs on two NVIDIA A100 GPUs until convergence.

\subsection{Evaluation Metrics}
\noindent{\bf MS-SSIM, LD, and AD.}
MS-SSIM~\cite{Wang_Bovik_Sheikh_Simoncelli_2004} is an image quality assessment method that measures structural similarity. It constructs a Gaussian pyramid to compute a weighted sum of SSIM across multiple scales, thereby mitigating the influence of sampling density. LD~\cite{You_Matsushita_Sinha_Bou_Ikeuchi_2018} quantifies distortion by evaluating the average local deformation at each pixel. This metric leverages SIFT Flow~\cite{liu2010sift} to align pixel positions and computes the mean L2 distance between corresponding pixels. AD~\cite{ma2022learning} serves as a robust metric for document dewarping by aligning an undistorted image with a reference image through translation and scaling, followed by distortion error computation.

\noindent{\bf ED and CER.}
The Edit Distance (ED)~\cite{lcvenshtcin1966binary} measures the minimum number of operations—deletion, insertion, and substitution—required to transform one string into another. The Character Error Rate (CER)~\cite{Morris_Maier_Green_2021} is derived from ED and is defined as $\mathrm{CER} = (d + s + r)/N$, where $N$ denotes the total number of characters in the reference string.

\subsection{Baseline Comparison}
\noindent{\bf Evaluation on the DocUNet.} 
As presented in Tab.~\ref{tbl:docunet},  we categorize methods into three groups: weakly supervised, foreground-based, and foreground-independent. Our distortion correction method surpasses existing approaches across almost all evaluated metrics. Notably, ForCenNet reduces the LD metric to 4.823. Compared to weakly supervised methods such as FDRNet~\cite{xue2022fourier} and PaperEdge~\cite{ma2022learning}, our method demonstrates superior performance in distortion metrics (MS-SSIM, LD, and AD). For fair comparison, we conduct an additional experiment on undistorted DOC3D dataset. ForCenNet-DOC3D lowers the ED metric below 600 for the first time, surpassing foreground-supervised methods such as FTDR~\cite{li2023foreground}, DocGeoNet~\cite{feng2022geometric}, and RDGR~\cite{jiang2022revisiting}. These findings underscore the significance of leveraging foreground features for improved distortion correction.

\begin{table}[t!]

\centering
\label{tbl:doc300}
\resizebox{\linewidth}{!}{
\begin{tabular}{llccccc}
\toprule
\textbf{Type} & \textbf{Model} & \textbf{MS-SSIM↑} & \textbf{LD↓} & \textbf{AD↓}  & \textbf{ED↓} & \textbf{CER↓} \\ 
\midrule
\multirow{1}{*}{WS.} & PaperEdge~\cite{ma2022learning}   & 0.583 & 8.00  & 0.255  & 704.34  & 0.221 \\ 
\midrule
\multirow{4}{*}{other.} & DewarpNet~\cite{das2019dewarpnet} & 0.492 & 13.94 & 0.331  & 1059.57 & 0.355 \\ 
 & DocTr~\cite{feng2021doctr}  & 0.616 & 7.21 & 0.254  & 699.63 & 0.223 \\ 
  & MetaDoc~\cite{dai2023matadoc}    & 0.638 & 5.75 & 0.178  & -- & -- \\
  & DocRes~\cite{zhang2024docres}   & 0.626  &6.83 & 0.241  & 774.80 & 0.241 \\
& CGU-Net~\cite{verhoeven2023uvdoc}  & 0.621  & 7.73  & 0.217 & 735.95 & 0.283 \\
\midrule
\multirow{5}{*}{FU.} & DDCP~\cite{xie2021document}  & 0.552 & 10.97 & 0.357  & 2130.01 & 0.552 \\ 
    & DocGeoNet~\cite{feng2022geometric}   & 0.638 & 6.40 & 0.242  & 664.96 & 0.218 \\ 
     & FTDR~\cite{li2023foreground}   & 0.607 & 7.68 & 0.244  & 652.80 & 0.211 \\
     & LA-DocFlatten~\cite{li2023layout}  & 0.651 & 5.70 & 0.195  & -- & -- \\
\rowcolor{myblue} 
{\cellcolor{fullwhite} } &
ForCenNet-DOC3D & \underline{0.709} & \underline{4.73} & \underline{0.136} & \underline{449.12} & \underline{0.153
} \\
\rowcolor{myblue} 
{\cellcolor{fullwhite} } &
ForCenNet & \textbf{0.713} & \textbf{4.65} & \textbf{0.123} & \textbf{390.61} & \textbf{0.138} \\

\bottomrule
\end{tabular}
}
\caption{\textbf{Result comparisons} between our proposed with existing methods on the DIR300 Benchmark~\cite{feng2022geometric}.}
\label{tab:DIR300}
\vspace{-5mm}
\end{table}

\begin{table*}[t!]
\centering

\resizebox{0.9\linewidth}{!}{
\begin{tabular}{lccccccccc} 
\toprule

\multirow{2}{*}{\textbf{Model}}  &  \multirow{2}{*}{\textbf{Pub.}} & 
\multicolumn{4}{c}{\textbf{WarpDoc}} & \multicolumn{4}{c}{\textbf{DocReal}}
\\

& & \makecell[c]{ \textbf{MS-SSIM$\uparrow$}} & \textbf{LD$\downarrow$} &  \textbf{AD$\downarrow$} & \textbf{ED$\downarrow$}  & \makecell[c]{ \textbf{MS-SSIM$\uparrow$}} &  \textbf{LD$\downarrow$} & \makecell[c]{ \textbf{AD$\downarrow$}} & \textbf{ED$\downarrow$} \\
\midrule
 DocTr~\cite{feng2021doctr} & ACM MM 21 &  0.39 &  27.01 &  0.77 & 1796.11 & 0.550 & 12.60 & 0.32  & 785.05 \\
 DocGeonet~\cite{feng2022geometric} & ECCV 22 &  0.40 &  24.71 &  0.75 & 1871.51&   0.553 & 12.23 & 0.31  &  784.47\\
 FDRNet~\cite{xue2022fourier} & CVPR 22 &  0.46 &  19.11 & -- & --  & -- & -- & --  & -- \\
 CGU-Net~\cite{verhoeven2023uvdoc} & SIGGRAPH 23 &  0.35 &  26.28 &  0.63 & 1760.84 & 0.549 & 11.33 & 0.27 &  753.35 \\
 DocReal~\cite{yu2024docreal} & WACV 23 & -- & -- & -- & --  & \underline{0.555} & \underline{9.82} & \underline{0.23}  &  \textbf{736.69}\\
 DocRes~\cite{zhang2024docres} & CVPR 24 &  \underline{0.5} &  \underline{12.86} &  \underline{0.45}  & \underline{1425.40} & 0.550 & 11.52 & 0.32  & 769.51 \\
\midrule
\belowrulesepcolor{myblue}
\rowcolor{myblue} ForCenNet  &  & \textbf{0.54} & \textbf{8.10} & \textbf{0.18} & \textbf{899.67}  & \textbf{0.595} & \textbf{6.95} & \textbf{0.17} & \underline{753.12} \\
\aboverulesepcolor{myblue}
\bottomrule
\end{tabular}}
\caption{\textbf{Generalization Comparison.} Columns 3-6 show the comparison results on WarpDoc~\cite{xue2022fourier}, while columns 7-10 present the results on DocReal~\cite{yu2024docreal}.}
\vspace{-3mm}
\label{tbl:warpdocdocreal}
\end{table*}

\noindent{\bf Evaluation on the DIR300.} As shown in Tab.~\ref{tab:DIR300}, our method outperforms state-of-the-art approaches, such as FTDR~\cite{li2023foreground}, on the DIR300 dataset. ForCenNet achieves an MS-SSIM score of 0.713, the highest reported to date, and reduces the LD metric to 4.653. In the OCR evaluation, following the protocols of DocGeoNet~\cite{feng2022geometric} and FTDR~\cite{li2023foreground}, we assess 90 images and reduce the ED metric to below 400, surpassing the previous leader, FTDR~\cite{li2023foreground}. These results underscore the effectiveness of our method in enhancing both distortion removal and OCR performance. We visualize the dewarping results on the DocUNet Benchmark~\cite{ma2018docunet} and DIR300 dataset~\cite{feng2022geometric}, comparing ForCenNet with existing methods (Fig.~\ref{fig:image7}). The initial examples demonstrate our method's ability to correct text regions with complex distortions, while the subsequent examples showcase the success of our foreground-centric approach in reducing distortions in table lines and intricate graphics.

\begin{table}[t]
\centering

\begin{tabular}{@{}llccccc@{}}
\toprule
 \textbf{Sample Size} & \textbf{MS-SSIM $\uparrow$} & \textbf{LD $\downarrow$} & \textbf{AD $\downarrow$} & \textbf{CER $\downarrow$} \\ 
\midrule
$\times1$   & 0.449 & 10.745 & 0.382 & 0.291 \\ 
$\times100$   & 0.530 & 5.348  & 0.231  &  0.208\\ 
$\times500$   & 0.566 & \textbf{4.892}  & 0.201  & 0.149 \\ 
$\times1000$   & \textbf{0.571} & 4.950  & \textbf{0.197}  & \textbf{0.141} \\
$\times2000$   & 0.567 & 4.965 & 0.203 & 0.147 \\
$\times5000$   & 0.569 & 4.942   & 0.209 & 0.151 \\
\bottomrule
\end{tabular}
\caption{\textbf{ForCenNet Performance on Different Dataset Sizes.} $\times$1000 indicates the application of 1,000 randomly generated deformation fields to the same image.}
\vspace{-3mm}
\label{tbl:table4}
\end{table}

\noindent{\bf Cross-domain Robustness. } 
Document correction models typically suffer from performance degradation in unseen environments with varying lighting conditions, viewpoints, and textures. Ensuring cross-domain robustness is crucial for real-world applications. To assess this property, we evaluate our method on two additional datasets, WarpDoc~\cite{xue2022fourier} and DocReal~\cite{yu2024docreal}, without incorporating undistorted reference images during continued training. The results in Tab.~\ref{tbl:warpdocdocreal} indicate that our method achieves competitive performance on previously unseen document images.

\subsection{Ablation Study}
We conduct ablation studies to evaluate the effectiveness of the proposed approach, focusing on label generation and core model architecture. All the ablation experiments are performed on the DocUNet~\cite{ma2018docunet} dataset.

\noindent{\bf Dataset Scaling.} We conduct an ablation study on dataset scaling using randomly generated sample sets. A total of 65 undistorted images from DocUNet~\cite{ma2018docunet} are used to construct experimental groups of varying sizes. Multiple distortion fields are applied to each image through label preprocessing. All groups are trained with identical epochs and hyperparameters. The results, summarized in Tab.~\ref{tbl:table4}, show that increasing dataset size improves model performance, highlighting the efficiency of the label generation module in iterative training with domain-specific undistorted images. However, performance gains plateau beyond a certain scale, indicating scalability limitations.

\begin{table}[t]
\centering
\resizebox{\linewidth}{!}{
\begin{tabular}{ccccccc}
\toprule
\textbf{ID} & \textbf{MGD} & \textbf{CL} & \textbf{Sample Size} &\makecell[c]{MS-SSIM$\uparrow$} & \textbf{LD $\downarrow$} & \textbf{CER $\downarrow$} \\
\midrule
A & \xmark   & \cmark & $\times 1000$ & 0.558 & 5.44 & 0.173 \\
B & \cmark &  \xmark & $\times 1000$ & 0.565 & 5.10 & 0.169 \\
C & \cmark & \cmark & $\times 1000$ & \textbf{0.571} & \textbf{4.95} & 
\textbf{0.141} \\
D & \xmark & \xmark & $\times 1000$ & 0.530 & 7.06 & 0.198\\
\bottomrule
\end{tabular}
}
\caption{\textbf{Ablation Study.} \textbf{MGD} denotes the Mask-Guided Transformer Decoder, and \textbf{CL} represents the Curvature Consistency Loss.}
\vspace{-3mm}
\label{abliation:table5}
\end{table}

\begin{figure}[t]

\centering\includegraphics[width=\linewidth]{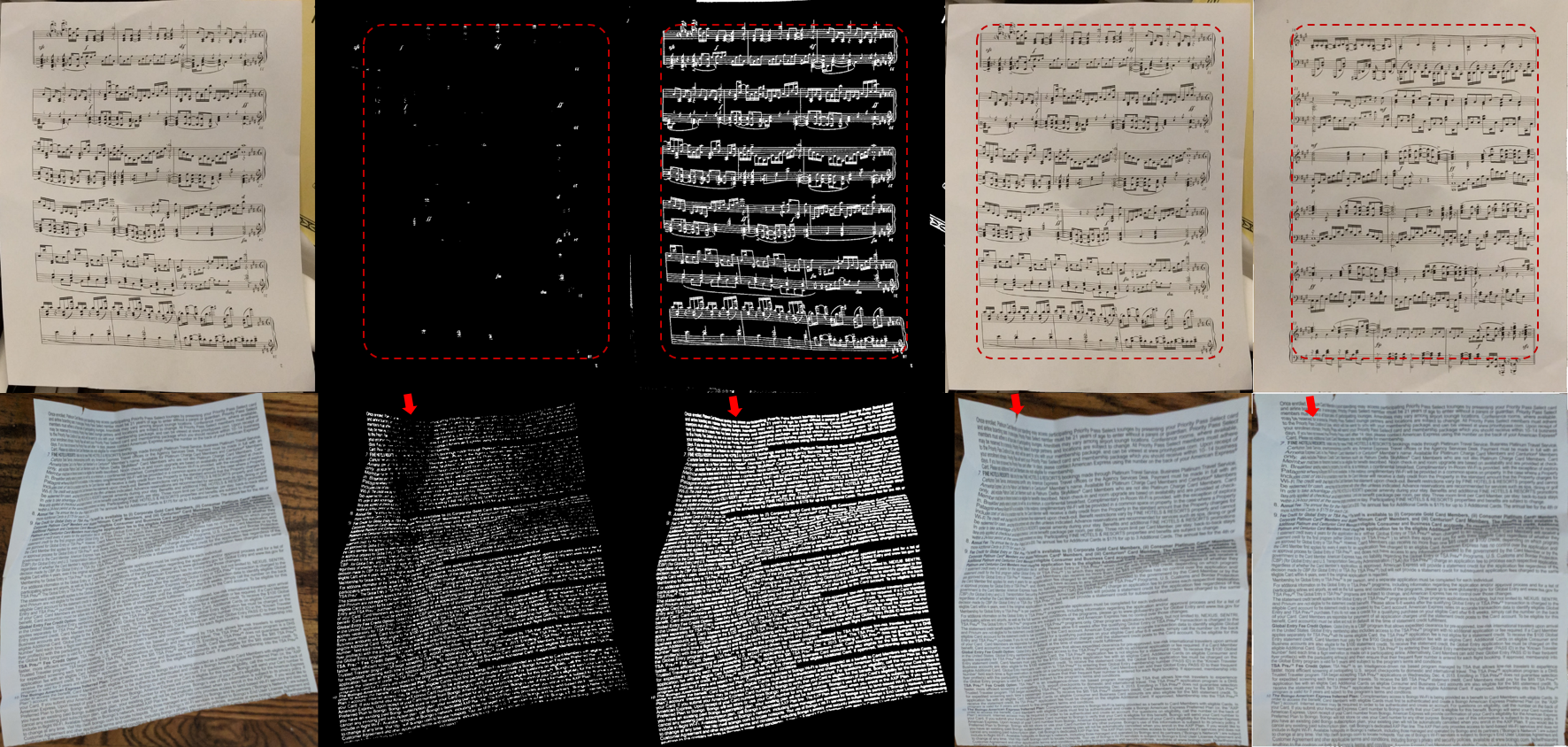}
\caption{\textbf{Visualization of Foreground Segmentation.} From left to right: the distorted input image, segmentation results from the frozen model and the differentiable model, followed by the corresponding dewarped outputs generated by each model.}
\vspace{-3mm}
\label{fig:image6}
\end{figure}

\noindent{\bf Structure Modifications.} We subsequently examine various components of the model, with the results summarized in Tab.~\ref{abliation:table5}. The comparison between experiments A and C reveals that incorporating the mask-guided module improves MS-SSIM from 0.558 to 0.571, while Local Distortion (LD) decreases from 5.44 to 4.95. Similarly, experiments B versus C show that removing curvature loss slightly reduces MS-SSIM from 0.571 to 0.565 and increases the Character Error Rate (CER) from 0.141 to 0.169. These findings indicate that generating sufficient distorted samples and effectively utilizing foreground information significantly enhance document image correction. By refining foreground definitions and emphasizing readable regions, dewarping performance is further improved.

\begin{figure}[t]
\centering\includegraphics[width=\linewidth]{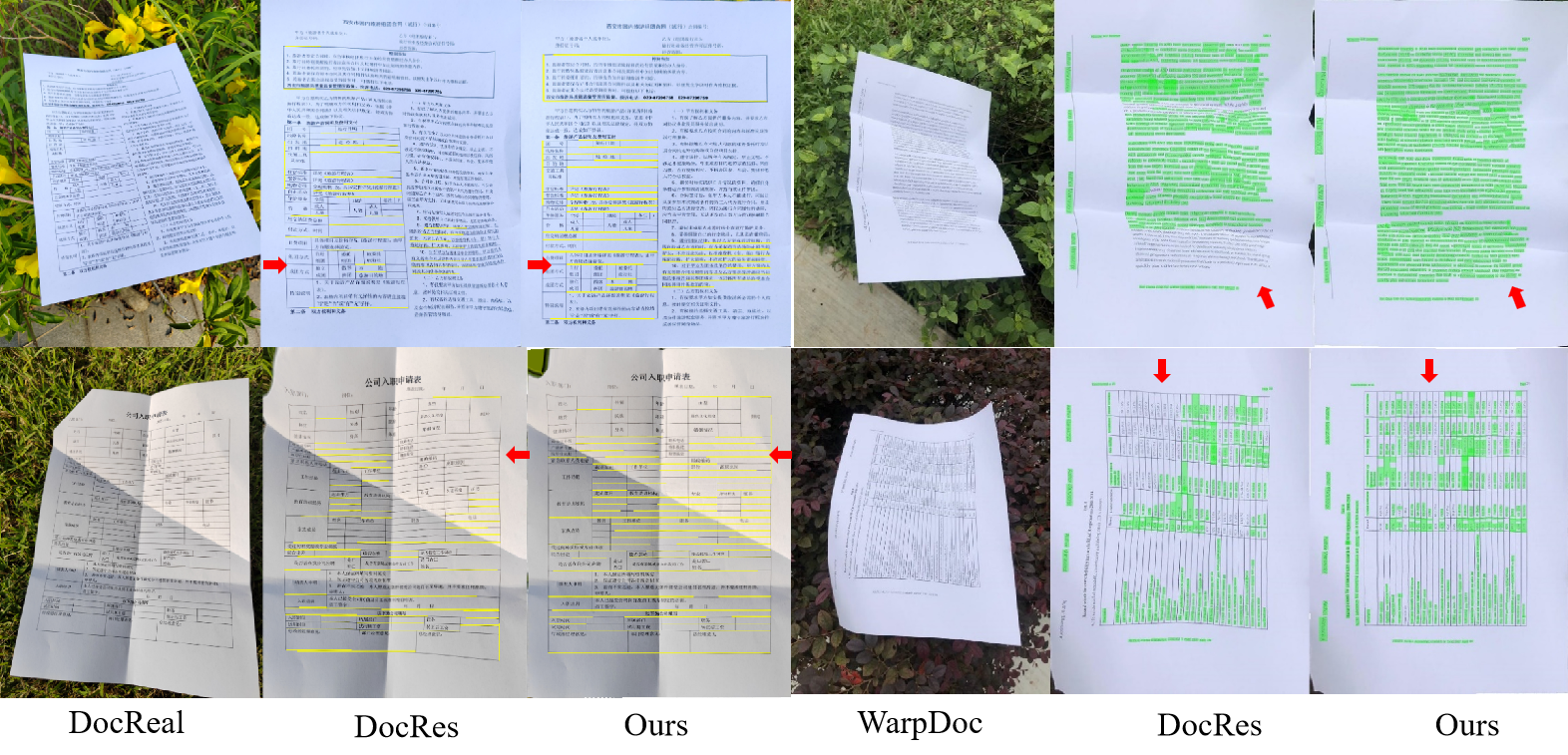}
\vspace{-3mm}
\caption{\textbf{Visualization of Foreground Results on WarpDoc~\cite{xue2022fourier} and DocReal~\cite{yu2024docreal} benchmarks. } Columns 1-3: detection results of line elements. Columns 4-6: detection results of text elements. Highlighted colors indicate detected regions, while red arrows mark differences.}
\vspace{-3mm}
\label{fig:image4}
\end{figure}

\noindent{\bf Ablation on Different Segmentation Models.} To assess the necessity of a differentiable foreground segmentation model, we replace it with a separately trained model and freeze its weights during dewarping training. Compared to the differentiable model, this modification leads to a rapid decline in MS-SSIM to 0.468 and an increase in Character Error Rate (CER) to 0.212. Fig.~\ref{fig:image6} presents two examples from DocUNet, illustrating that rare characters and complex distortions severely impact foreground segmentation performance, consequently degrading the final dewarping results.

\begin{figure}[t]
\centering\includegraphics[width=\linewidth]{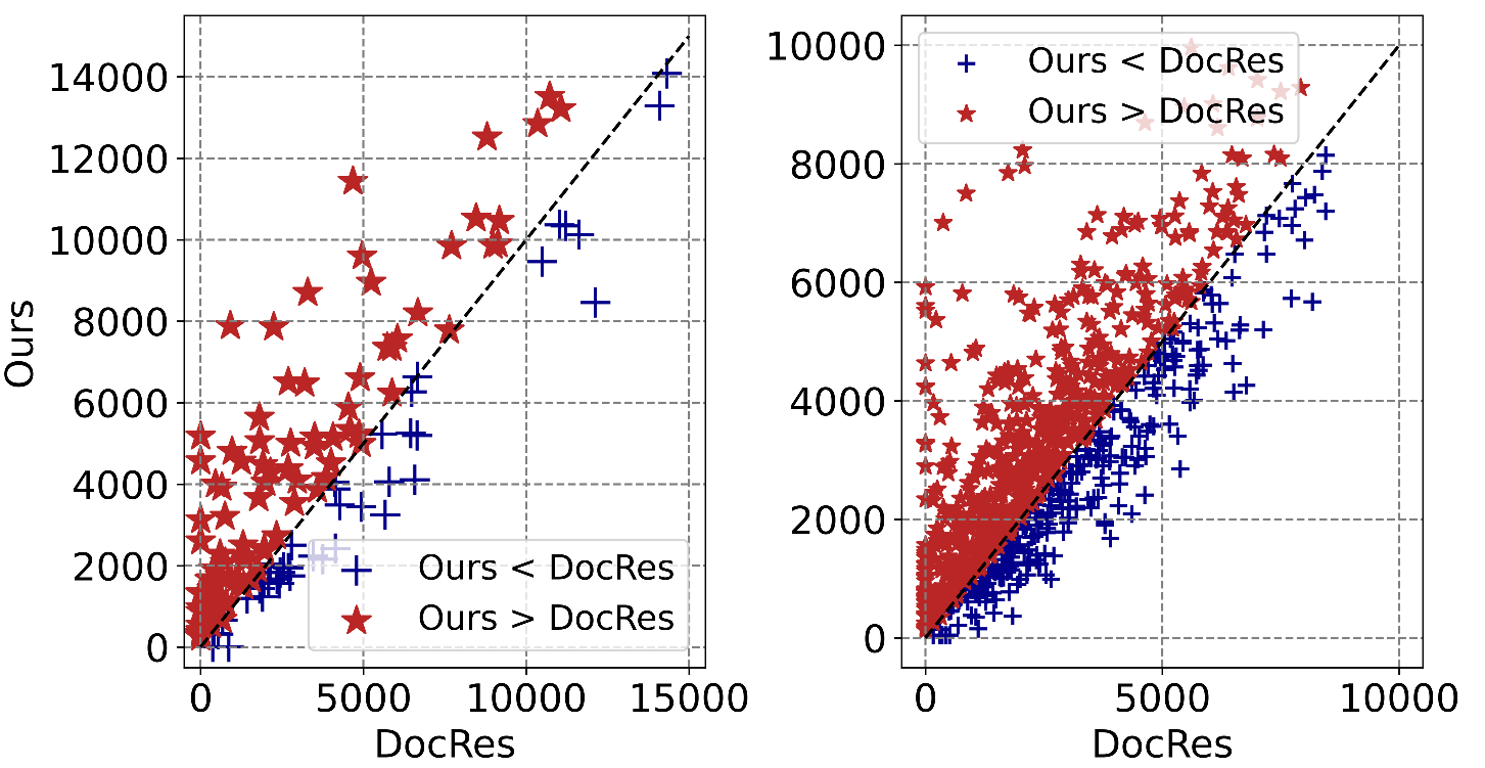}
\vspace{-3mm}
\caption{\textbf{Quantitative evaluation of straight-line rectification.} The first image displays results on the DocReal~\cite{yu2024docreal} dataset, while the second image presents results on the WarpDoc~\cite{xue2022fourier} dataset.}
\vspace{-3mm}
\label{fig:image12}
\end{figure}

\subsection{Experimental Analysis}

\noindent{\bf Visualization of Foreground Results.}
We visualize foreground detection in distorted images. Fig.~\ref{fig:image4} compares the performance of DocRes~\cite{zhang2024docres} and our method using the Tesseract OCR engine (v5.0.1)~\cite{smith2007overview} and the line detection results produced by Algorithm~\ref{algorithm}. The dewarping process significantly improves the recall of text and line elements. To quantify the correction capability of our model for straight line elements, we detect straight lines in the corrected images and calculate the total line length for each sample. We compare our method with DocRes on the WarpDoc~\cite{xue2022fourier} and DocReal~\cite{yu2024docreal} datasets, as shown in Fig.~\ref{fig:image12}. Our method outperforms DocRes on 65\% of DocReal samples and 69\% of WarpDoc samples. For challenging cases, unclear boundaries between foreground and background slightly reduce segmentation accuracy. Overall, guided by foreground masks and curvature supervision, our model significantly enhances geometric perception and image readability.

\begin{figure}[t]
\centering\includegraphics[width=\linewidth]{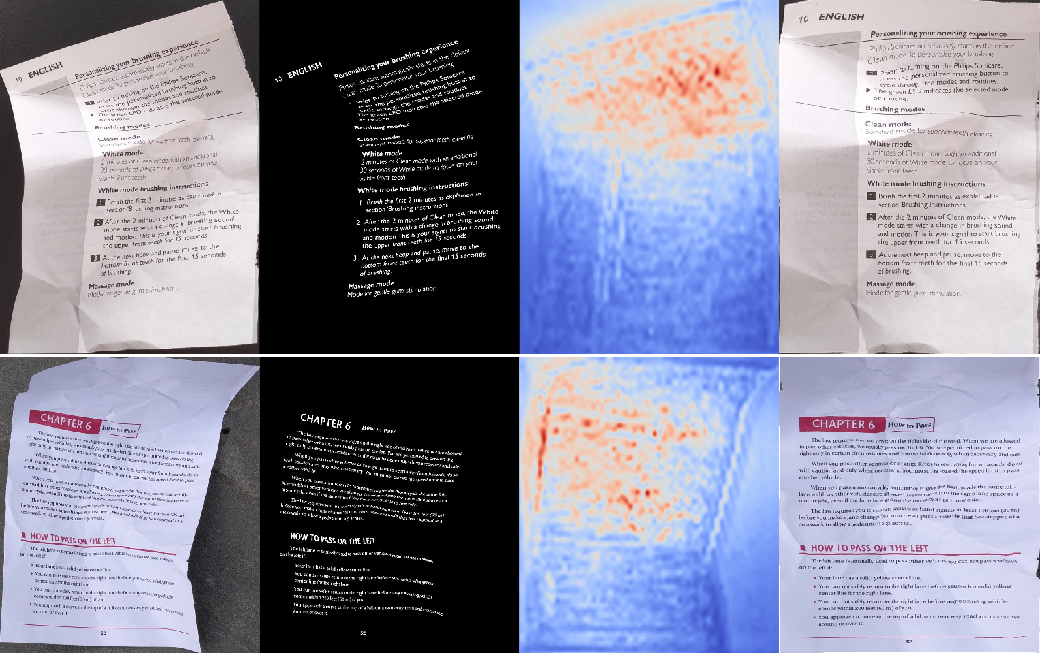}
\vspace{-3mm}
\caption{\textbf{Visualization of intermediate layer results.} Left to right: distorted image, foreground mask, foreground attention map, and rectification results of our model.}
\vspace{-3mm}
\label{fig:image5}
\end{figure}

\noindent{\bf Intermediate Results Visualization.} 
To improve model interpretability, we visualize the foreground segmentation results and the attention maps from the decoder’s final layer. Fig.~\ref{fig:image5} demonstrates that the differentiable mask prediction module effectively handles complex distortions, such as folds, creases, and shadows. Concurrently, the heatmap distribution in the attention maps demonstrates the model's ability to identify distorted regions. Overall, our model successfully captures the foreground mask, guiding subsequent modules to focus on the distorted areas within the foreground, thereby facilitating the reconstruction of an undistorted document image.

\par

\section{Conclusion}
\indent This paper explores the use of naturally occurring foreground elements in documents for rectification. We propose ForCenNet, a framework that enables efficient model training using only undistorted images. Our method extracts foreground labels directly from undistorted images, leveraging a mask-guided module to enhance the model’s focus on foreground information. Furthermore, we introduce a curvature consistency loss to improve the model’s geometric understanding of fine structures such as lines and text. Extensive experiments on four public benchmark datasets demonstrate the effectiveness of our approach. In future work, as ForCenNet can generate mask predictions, we will investigate its integration with image scanning and enhancement.

{
    \small
    \bibliographystyle{ieeenat_fullname}
    \bibliography{main}
}

\clearpage
\setcounter{page}{1}
\maketitlesupplementary

\section{Experimental Details}
In the Tab.\ref{tbl:table6}, we present the detailed experimental setup and model hyperparameters. To investigate the impact of sampling density on geometric fidelity, we evaluate line sampling intervals of 2, 4, 8, 16, and 32 pixels on the DocUNet dataset. The corresponding MS-SSIM scores are 0.582, 0.582, 0.579, 0.575, and 0.569, respectively. These results indicate that finer sampling better captures local curvature, thereby enhancing perceptual quality. Considering both reconstruction performance and computational efficiency, we adopt a sampling interval of 4 pixels as a trade-off. To bridge the domain gap between synthetic and real-world data, we synthesize more realistic training samples by compositing geometrically distorted document images onto randomly selected COCO17 backgrounds. Rather than explicitly modeling illumination artifacts such as shadows, we leverage standard data augmentation strategies to implicitly simulate such effects, enhancing the model’s generalization to diverse visual contexts.

\begin{table}[h]
\centering

\resizebox{0.3\textwidth}{!}{
\begin{tabular}{@{}l|c}
\toprule
Hyperparameter & Value \\
\midrule

Learning rate & 0.0001 \\
Scaling factor $\sigma$ & 0.005 \\
Positive value $\varepsilon$ & 0.0001 \\
Smoothing coefficient $\gamma$ & 0.8\\
Batch size & 32 \\
Warm-up phase & 0.2 \\
Training epochs & 30 \\
GPU & $2\times A100$ \\
\bottomrule
\end{tabular}
}
\caption{The model hyperparameters}
\label{tbl:table6}
\vspace{-5mm}
\end{table}

\section{Data augmentation}
\begin{figure}[h]
\centering\includegraphics[width=0.5\textwidth]{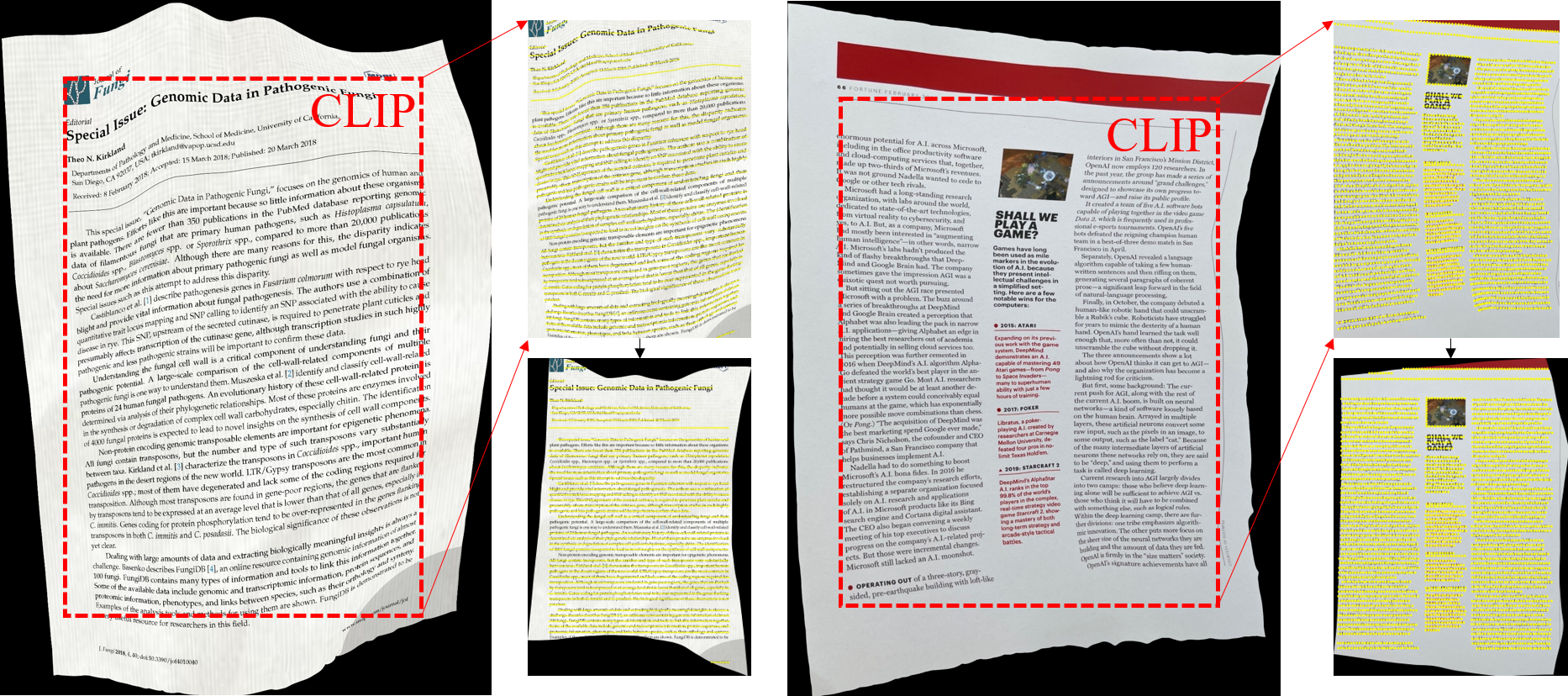}
\caption{\textbf{Visualization of the cropping process}}

\label{fig:image15}
\end{figure}
To enhance the diversity of distortion relationships, we apply minor cropping to the generated distorted images and compute the correspondences of the deformation field before and after cropping. Fig.~\ref{fig:image15} shows the cropped distorted images along with their corresponding rectified document images. Additionally, we process the foreground lines and masks required by ForCenNet and highlight them in yellow.

\section{Downstream tasks}
\begin{figure}[h]
\centering\includegraphics[width=0.5\textwidth]{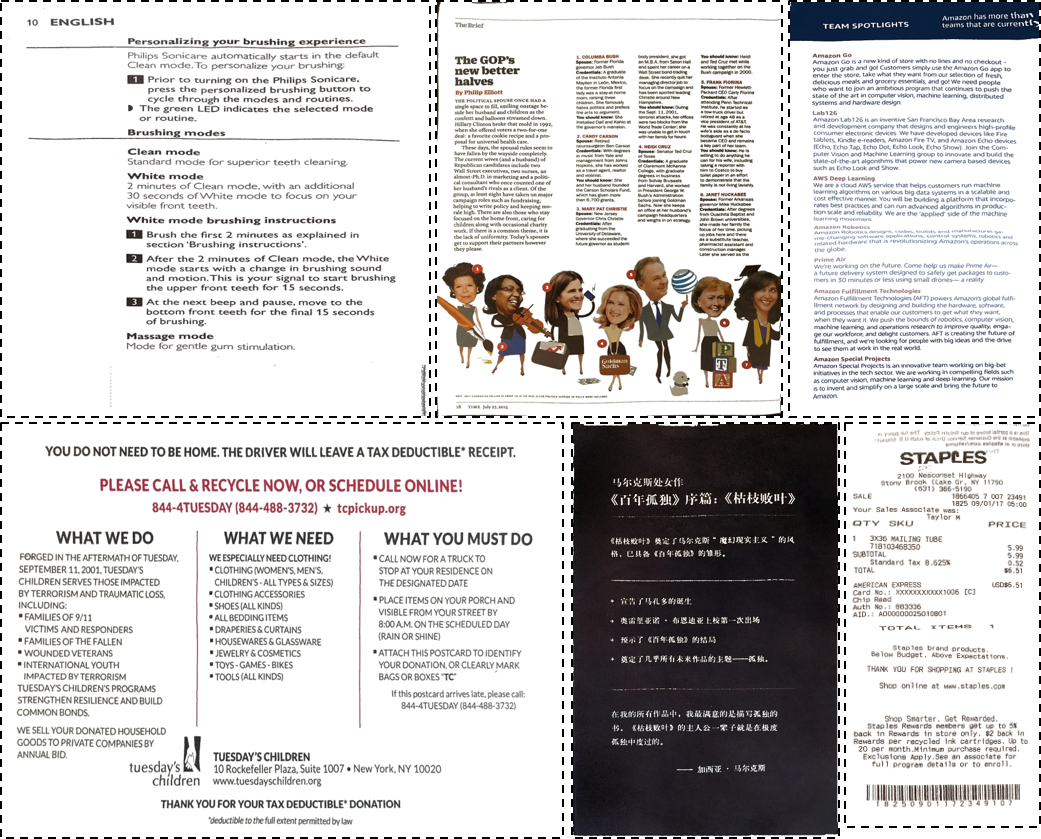}
\caption{\textbf{Exploration of enhancement tasks.}}
\vspace{-5mm}
\label{fig:image16}
\end{figure}

We conduct an exploratory investigation into a significant downstream task, namely appearance enhancement, also known as illumination correction. This task aims to restore a pristine appearance similar to that produced by a scanner or digital PDF file, without being limited to specific degradation types. In our study, we leverage the model-predicted foreground mask to assign non-foreground regions to white, while preserving the original colors of the foreground regions, simulating document enhancement effects. The selected visualization results from the DocUNet dataset are shown in Fig.~\ref{fig:image16}. Furthermore, we quantitatively evaluate the enhanced images against the ground-truth enhanced images from the DocUNet dataset, achieving an MS-SSIM score of 0.6712. These results highlight the potential of the ForCenNet architecture in synthetic enhancement tasks.

\section{Bias Analysis }
In the label preprocessing module, the forward map ($\mathcal{FM}$) is derived by proportionally sampling anchor points on the backward map ($\mathcal{BM}$) and constructing an augmentation matrix. However, a certain bias arises due to incomplete sampling. To quantify this bias, we apply two mappings ($\mathcal{FM}$ and then $\mathcal{BM}$) sequentially to the original image, line elements, and foreground masks. For the original image $I_u$, we compute SSIM with the mapped image. For line elements $L_u$, we calculate the displacement of points after mapping. For foreground elements $M_u$, we compute the IoU of the masks. The bias is quantified using 100,000 backward maps from DOC3D \cite{das2019dewarpnet}. The specific calculations are as follows:
\begin{equation}
\begin{aligned}
\label{eq:tilde_M}
Bias(I_u) &= SSIM(\mathcal{BM} (\mathcal{FM}(I_u)),I_u), \\
Bias(L_u) &= OFFSET(\mathcal{BM} (\mathcal{FM}(L_u)),L_u), \\
Bias(M_u) &= IOU(\mathcal{BM} (\mathcal{FM}(M_u)),M_u).  \\
\end{aligned}
\end{equation}\par
\begin{figure}[ht]
\centering\includegraphics[width=0.48\textwidth]{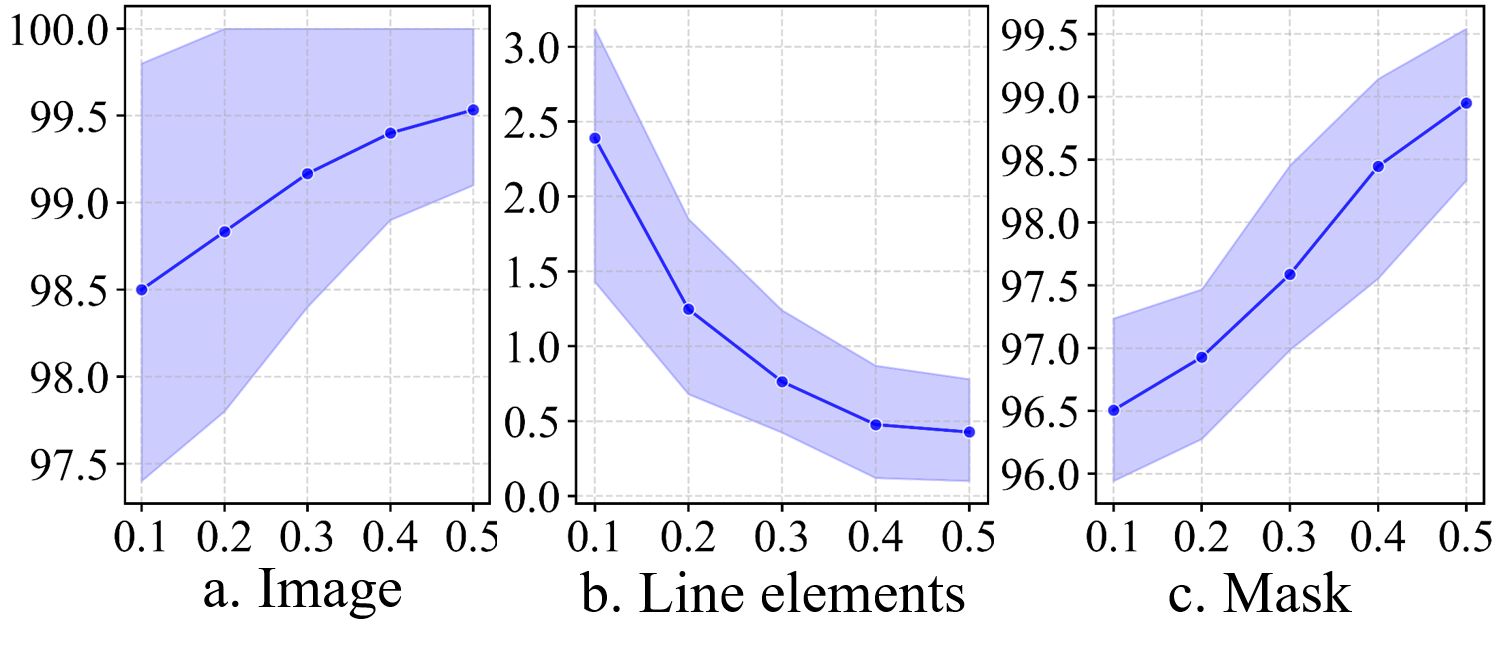}
\caption{\textbf{Statistic analysis of bias.}}
\vspace{-5mm}
\label{fig:image11}
\end{figure}

Taking line elements as an example, we calculate the displacement before and after the two mappings for all 100,000 samples, then compute the minimum, maximum, and average displacements. Fig. \ref{fig:image11} shows the displacement variation under different sampling ratios. As the sampling ratio increases, the displacement is effectively controlled. By balancing these metrics, we determine that a 40\% sampling ratio is optimal.

\section{More visualizations}
\label{sec:add_vis}
\begin{figure*}[ht]
\centering\includegraphics[width=0.95\linewidth]{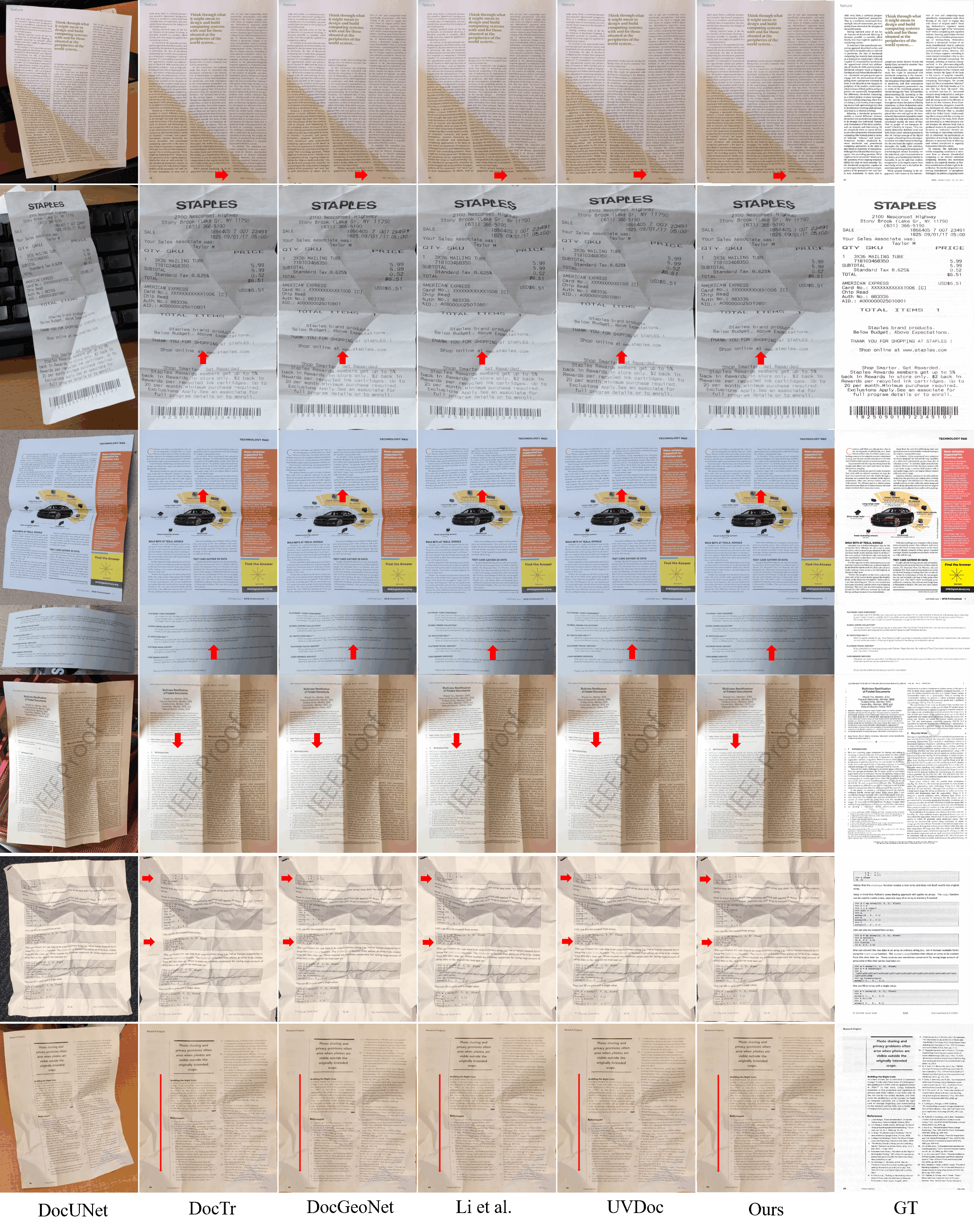}
\caption{\textbf{Visualization comparison on the DocUNet dataset.} }\label{fig:image17}
\end{figure*}

\begin{figure*}[ht]
\centering\includegraphics[width=0.95\linewidth]{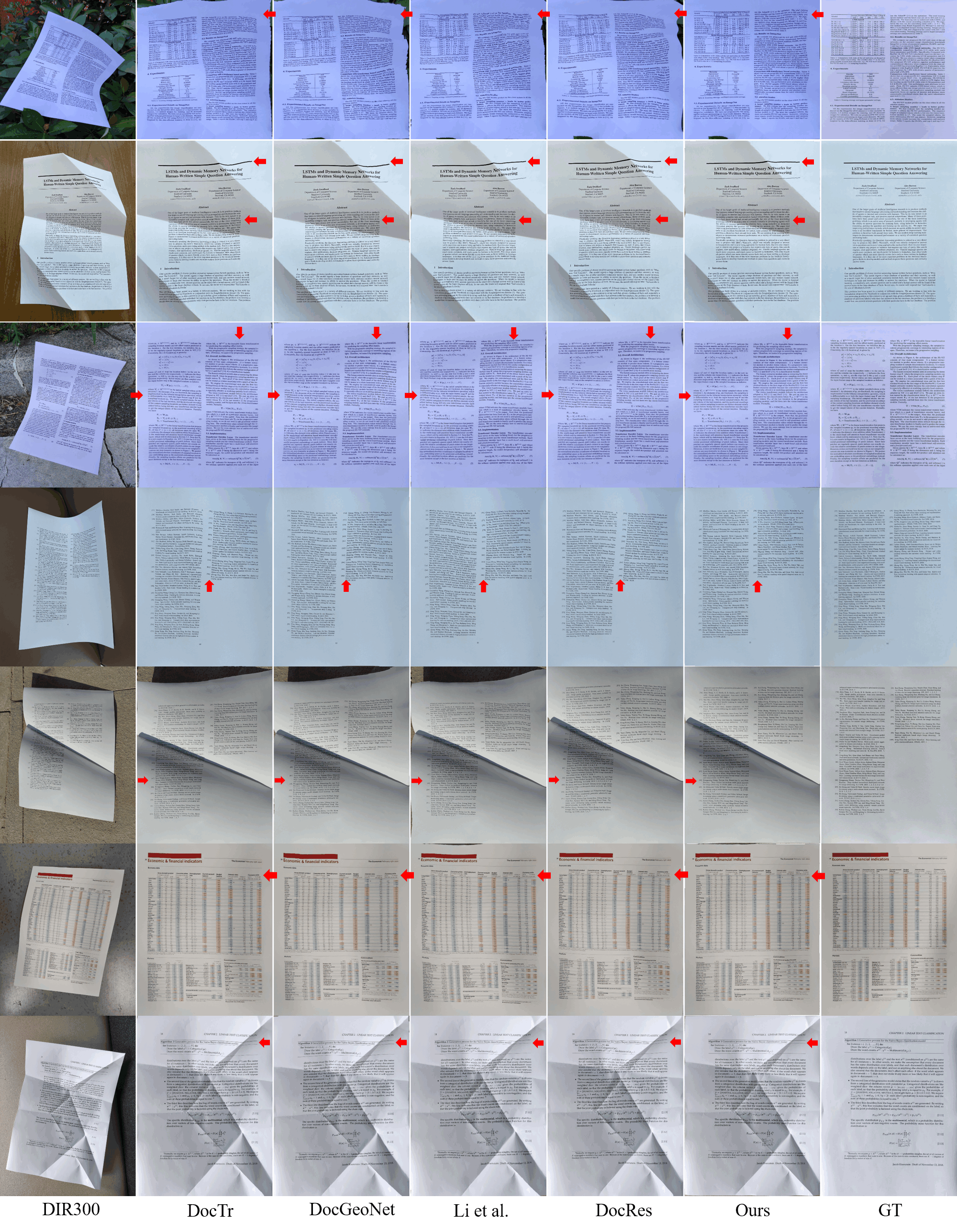}
\caption{\textbf{Visualization comparison on the DIR300 dataset.} }\label{fig:image18}
\end{figure*}

\begin{figure*}[ht]
\centering\includegraphics[width=0.95\linewidth]{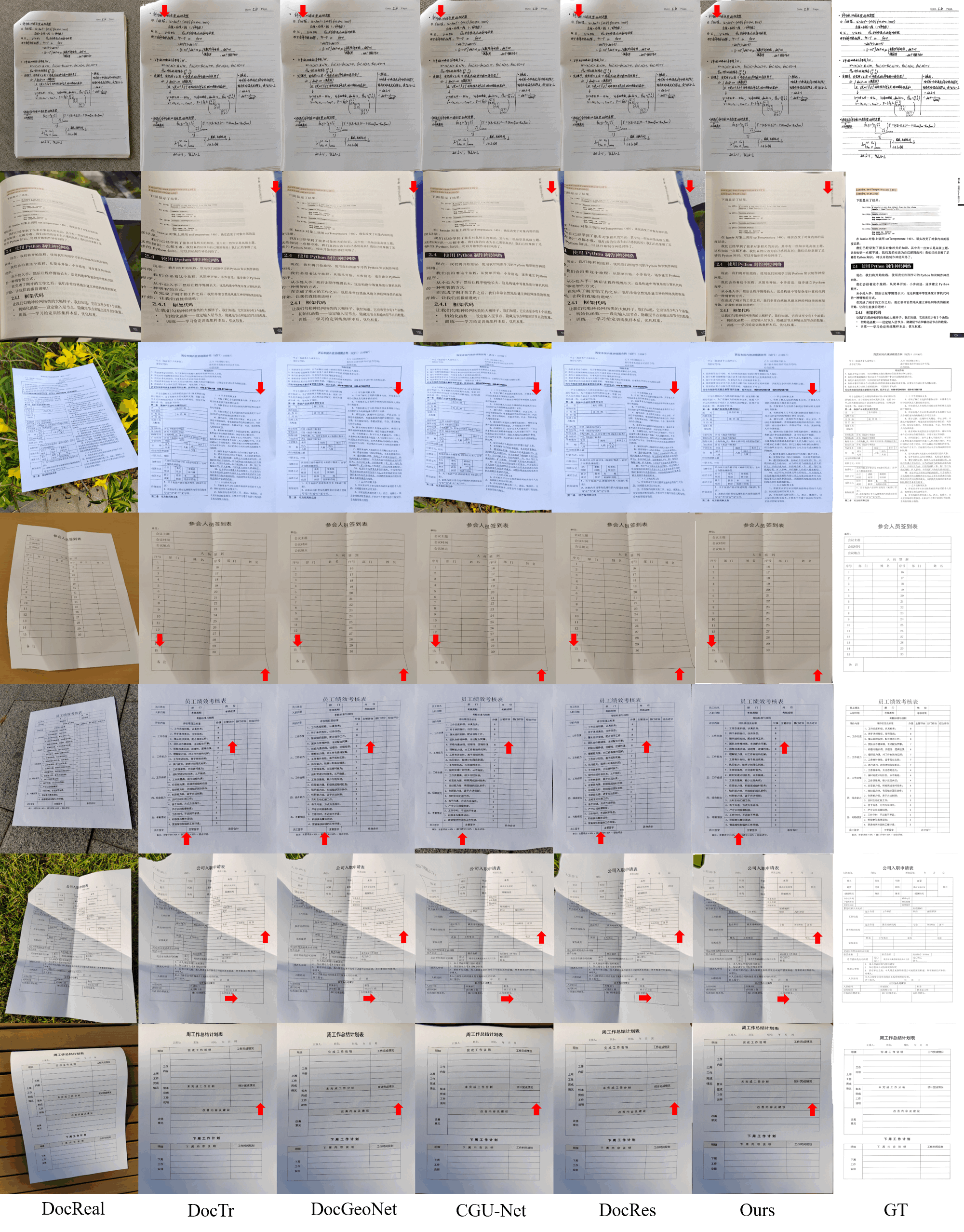}
\caption{\textbf{Visualization comparison on the DocReal dataset.} }\label{fig:image19}
\end{figure*}

\begin{figure*}[ht]
\centering\includegraphics[width=0.95\linewidth]{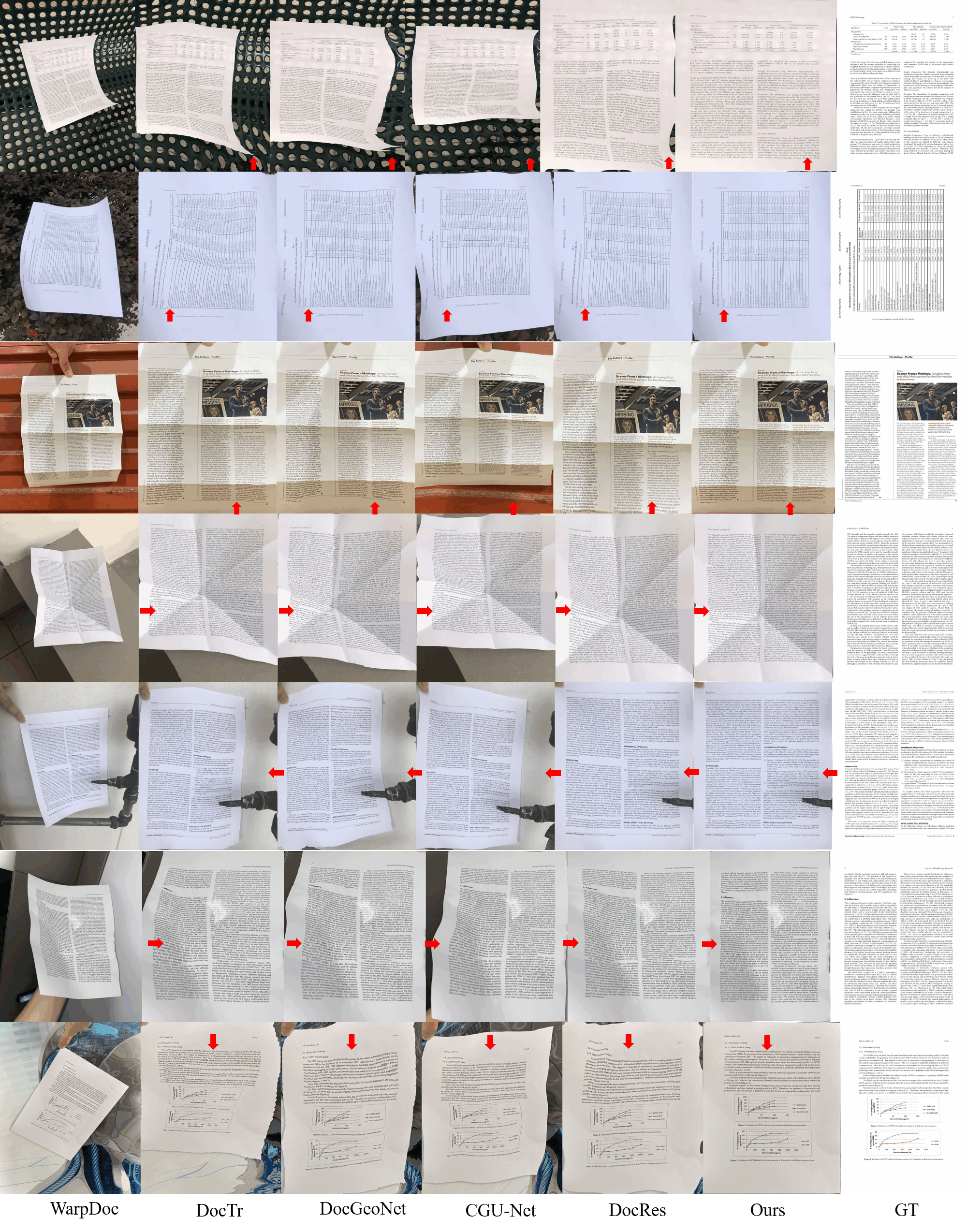}
\caption{\textbf{Visualization comparison on the WarpDoc dataset.} }\label{fig:image20}
\end{figure*}

We present additional comparisons of the model’s dewarping results in Fig.~\ref{fig:image17} , \ref{fig:image18} ,\ref{fig:image19} and \ref{fig:image20}, which effectively demonstrate the superiority of our approach. Red arrows highlight the differences.

\end{document}